%% file: main.tex
\documentclass[runningheads]{llncs}
\usepackage{graphicx}
\usepackage{array}
\usepackage{booktabs}
\usepackage{tikz}
\usepackage{standalone}
\usepackage{amsmath}
\usepackage{amssymb}
\usepackage{subfig}
\usepackage{hyperref}
\begin{document}
\bibliographystyle{splncs04}
\title{Structured State Space Models for Multiple Instance Learning in Digital Pathology}
\titlerunning{State Space Models in Digital Pathology}

\author{Leo Fillioux\thanks{These authors contributed equally to this work.}\and
Joseph Boyd$^\star$ \and
Maria Vakalopoulou\and
Paul-Henry Cournède\and
Stergios Christodoulidis}

\authorrunning{L. Fillioux et al.}

\institute{MICS Laboratory, CentraleSup\'elec, Universit\'e Paris-Saclay,
91190 Gif-sur-Yvette, France
\email{firstname.lastname@centralesupelec.fr}}

% \author{Anonymous submission}
% %
% \authorrunning{Anonymous}
% \institute{Anonymous institute}
% % \institute{Princeton University, Princeton NJ 08544, USA \and
% % Springer Heidelberg, Tiergartenstr. 17, 69121 Heidelberg, Germany
% % \email{lncs@springer.com}\\
% % \url{http://www.springer.com/gp/computer-science/lncs} \and
% % ABC Institute, Rupert-Karls-University Heidelberg, Heidelberg, Germany\\
% % \email{\{abc,lncs\}@uni-heidelberg.de}}
% %
\maketitle              % typeset the header of the contribution
\begin{abstract}
Multiple instance learning is an ideal mode of analysis for histopathology data, where vast whole slide images are typically annotated with a single global label. In such cases, a whole slide image is modelled as a collection of tissue patches to be aggregated and classified. Common models for performing this classification include recurrent neural networks and transformers. Although powerful compression algorithms, such as deep pre-trained neural networks, are used to reduce the dimensionality of each patch, the sequences arising from whole slide images remain excessively long, routinely containing tens of thousands of patches. Structured state space models are an emerging alternative for sequence modelling, specifically designed for the efficient modelling of long sequences. These models invoke an optimal projection of an input sequence into memory units that compress the entire sequence. In this paper, we propose the use of state space models as a multiple instance learner to a variety of problems in digital pathology. Across experiments in metastasis detection, cancer subtyping, mutation classification, and multitask learning, we demonstrate the competitiveness of this new class of models with existing state of the art approaches. Our code is available at \href{https://github.com/MICS-Lab/s4\_digital\_pathology}{https://github.com/MICS-Lab/s4\_digital\_pathology}.
\end{abstract}
\section{Introduction}
\input{sections/1_introduction}

\section{Related work}
\input{sections/2_related_work}

\section{Method}
\input{sections/3_methods}

\section{Experiments and discussion}
\input{sections/4_results}

\section{Conclusions}
\input{sections/5_conclusions}

\bibliography{IEEEabrv,3313}
\appendix
\section{Supplementary material}
\input{sections/6_supplementary}

\end{document}

% \section*{Supplementary materials}

\title{Structured State Space Models for Multiple Instance Learning in Digital Pathology (Supplementary Material)}
\titlerunning{State Space Models in Digital Pathology}

\author{Leo Fillioux\thanks{These authors contributed equally to this work.}\and
Joseph Boyd$^*$ \and
Maria Vakalopoulou\and
Paul-Henry Cournède\and
Stergios Christodoulidis}
%index{Boyd, Joseph}
%index{Villar, Irène}
%index{Mathieu, Marie-Christine}
%index{Deutsch, Eric}
%index{Paragios, Nikos}
%index{Vakalopoulou, Maria}
%index{Christodoulidis, Stergios}

\authorrunning{L. Fillioux et al.}

\institute{MICS Laboratory, CentraleSup\'elec, Universit\'e Paris-Saclay,
91190 Gif-sur-Yvette, France}
% \email{firstname.lastname@centralesupelec.fr}}

% \author{Anonymous submission}
% %
% \authorrunning{Anonymous}
% \institute{Anonymous institute}
% % \institute{Princeton University, Princeton NJ 08544, USA \and
% % Springer Heidelberg, Tiergartenstr. 17, 69121 Heidelberg, Germany
% % \email{lncs@springer.com}\\
% % \url{http://www.springer.com/gp/computer-science/lncs} \and
% % ABC Institute, Rupert-Karls-University Heidelberg, Heidelberg, Germany\\
% % \email{\{abc,lncs\}@uni-heidelberg.de}}
% %
\maketitle              % typeset the header of the contribution

\begin{figure}
\centering
\subfloat[TransMIL]{\includegraphics[width=0.28\textwidth]{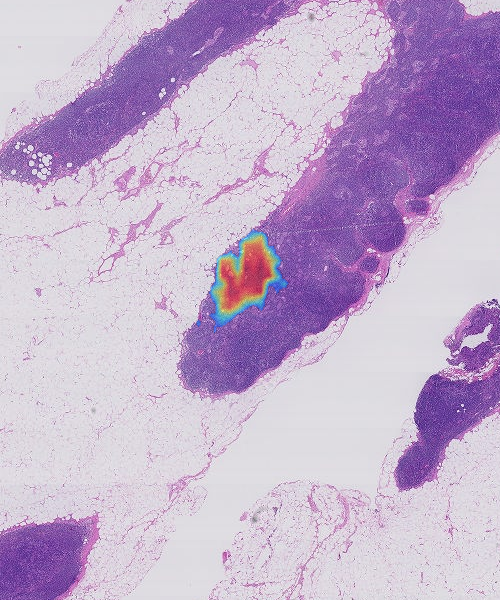}}\label{subfig:ihcsamples}
\subfloat[Ours]{\includegraphics[width=0.28\textwidth]{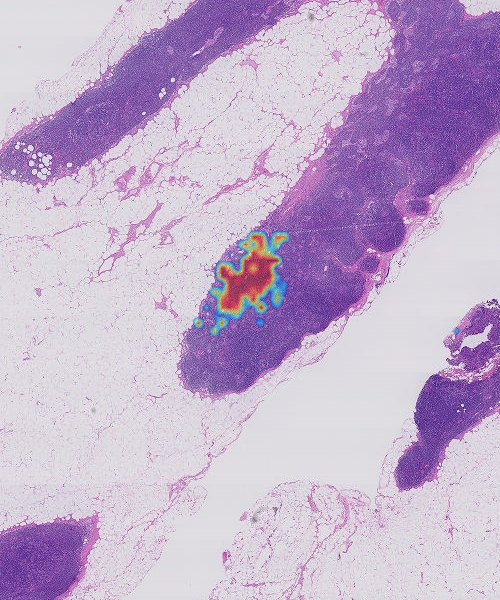}}\label{subfig:ihcsamples}
\subfloat[Ground truth mask]{\includegraphics[width=0.28\textwidth]{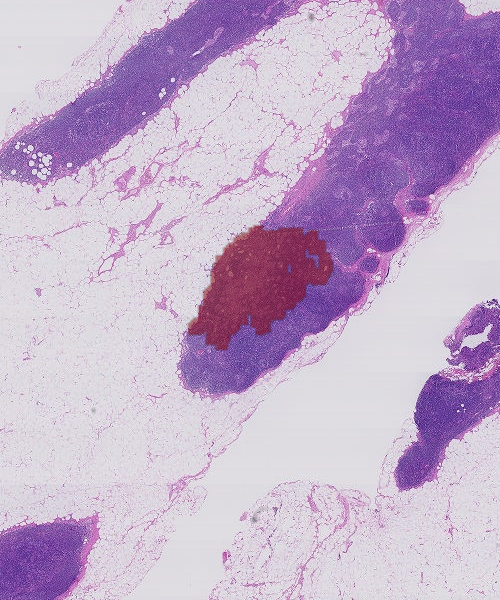}}\label{subfig:hessamples} 
\caption{Heatmap visualisations of metastatic region after multitask training on CAMELYON16 for TransMIL (a), proposed model (b), and ground truth (c).}
\label{fig:heatmaps}
\end{figure}

\begin{figure}
\centering
\includegraphics[width=0.85\textwidth]{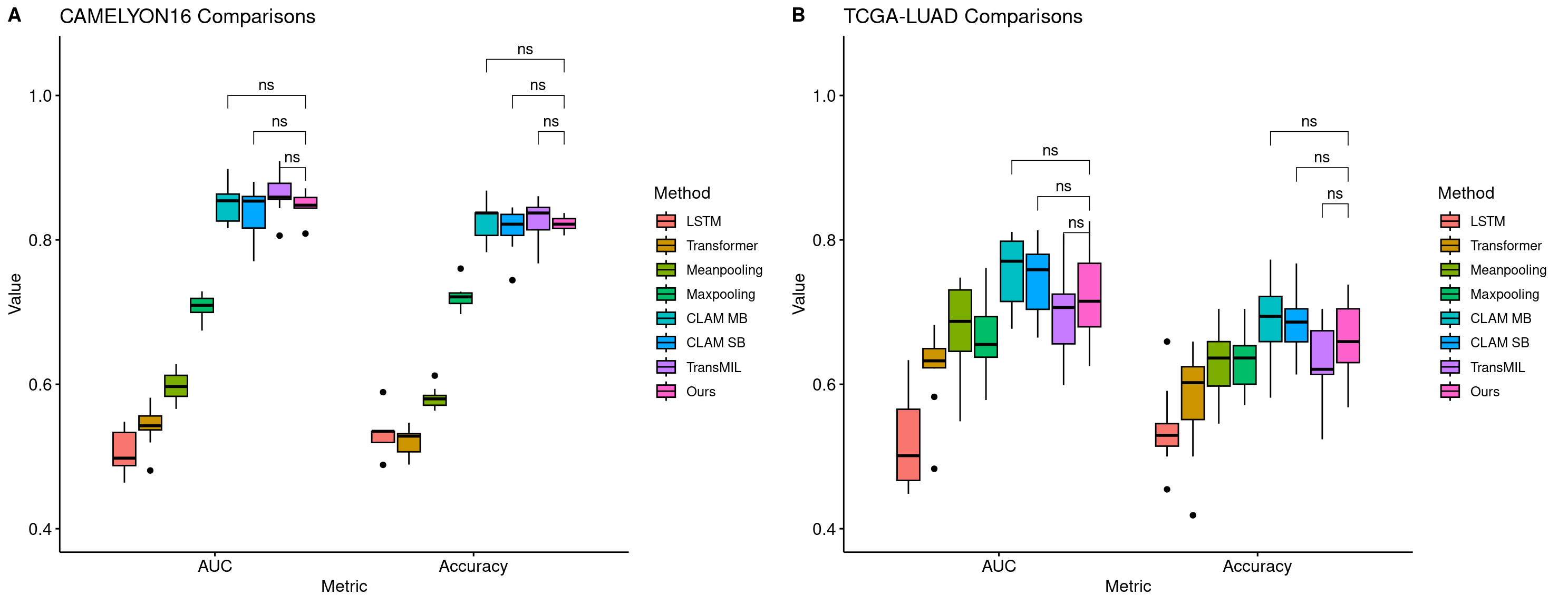}
\caption{AUC and accuracy over 10 folds for all methods on the CAMELYON16 dataset (A) and the TCGA-LUAD dataset (B). Significance estimated with pairwise $t$-tests (confidence level of 0.95)}
\label{fig:suppl_1}
\end{figure}

%% file: sections/1_introduction.tex
% * Some basic points on histo analysis e.g. MIL is useful because of sparsely available annotations
% * A high-level overview of the paper e.g. "in section 2 we do XXX, in section 3 we do YYY" etc.

% Cancer is the leading cause of death worldwide according to the World Health Organization. Recently, precision medicine efforts are shifting cancer care standards by providing novel personalised treatment plans with promising outcomes. Patient selection for such treatment regimes is based principally on the assessment of tissue biopsies and the characterisation of the tumor microenvironment. This is typically performed by experienced pathologists, who closely inspect chemically stained histopathological whole slide images (WSIs). Among the different staining protocols, the most commonly used is hematoxylin and eosin staining (H\&E).

%The most common tissue staining that it is routinely used is hematoxylin and eosin staining (H\&E). Under this staining the structural characteristics of the tissue are highlighted such as the nuclei and stroma. Alternatively, different staining protocols can be adopted that highlight specific cell characteristics like the expression of HER2, TP53, Ki67 and others. However, these are not always available and increase the cost and time for a more informed decision.

Precision medicine efforts are shifting cancer care standards by providing novel personalised treatment plans with promising outcomes. Patient selection for such treatment regimes is based principally on the assessment of tissue biopsies and the characterisation of the tumor microenvironment. This is typically performed by experienced pathologists, who closely inspect chemically stained histopathological whole slide images (WSIs). Increasingly, clinical centers are investing in the digitisation of such tissue slides to enable both automatic processing as well as research studies to elucidate the underlying biological processes of cancer. The resulting images are of gigapixel size, rendering their computational analysis challenging. To deal with this issue, multiple instance learning (MIL) schemes based on weakly supervised training are used for WSI classification tasks. In such schemes, the WSI is typically divided into a grid of patches, with general purpose features derived from pretrained ImageNet~\cite{russakovsky2015imagenet} networks extracted for each patch. These representations are subsequently pooled together using different aggregation functions and attention-based operators for a final slide-level prediction.

% Among the different aggregation functions, sequential models such as long short-term memory (LSTM) networks have been proven to work well on MIL settings both for visual cognition tasks~\cite{wang2020defense} as well as computational pathology~\cite{agarwalla2017representation}. However, in this work, we argue that recent sequential models such as the structure state space models ~\cite{gu2021efficiently} could provide faster and more efficient alternatives for the challenging tasks of gigapixel WSI classification.

State space models are designed to efficiently model long sequences, such as the sequences of patches that arise in WSI MIL. In this paper, we present the first use of state space models for WSI MIL. Extensive experiments on three publicly available datasets show the potential of such models for the processing of gigapixel-sized images, under both weakly and multi-task schemes. Moreover, comparisons with other commonly used MIL schemes highlight their robust performance, while we demonstrate empirically  the superiority of state space models in processing the longest of WSI sequences with respect to commonly used MIL methods.

%% file: sections/2_related_work.tex
% * Some basic points on MIL modeling
% * Some basic points on S4 models and how they compare with RNNs, transformers

% Sequence modeling

Using pretrained networks for patch-wise feature extraction is a well established strategy for histopathology analysis~\cite{coudray2018classification,tellez2019neural}. An extension of this approach is with MIL, where the patch-wise features of an entire slide are digested simultaneously by an aggregator model, such as attention-based models CLAM~\cite{lu2021data} and TransMIL~\cite{shao2021transmil}, the latter being a variant of self-attention transformers~\cite{vaswani2017attention}. \cite{chen2022scaling} proposes another transformer-based method in the form of a hierarchical ViT. Similar to our multitask experiments, \cite{gao2023semi} explores combining slide-level and tile-level annotations with a minimal point-based annotation strategy. One of the key components of MIL methods is the aggregation module that pools together the set of patch representations. Mean or max pooling operations are among the simplest and most effective for aggregating predictions over a whole slide~\cite{campanella2019clinical}. In contrast, recurrent neural networks (RNN) with long short-term memory (LSTM)~\cite{hochreiter1997long} model the patches more explicitly as a set of tokens in sequence. In particular, LSTM networks have been shown to work well in different MIL settings including both visual cognition~\cite{wang2020defense} and computational pathology~\cite{agarwalla2017representation}.

The state space model is a linear differential equation,

\begin{equation}
\label{eq:ss}
\begin{aligned}
\dot{x}(t) &= \boldsymbol{A}x(t) + \boldsymbol{B}u(t) \\ 
y(t) &= \boldsymbol{C}x(t) + \boldsymbol{D}u(t)
\end{aligned}
\end{equation}

that is widely studied in control theory, and describes a continuous time process for input and output signals $u(t) \in \mathbb{R}^p$ and $y(t) \in \mathbb{R}^q$, and state signal $x(t) \in \mathbb{R}^n$, and where the process is governed by matrices $\boldsymbol{A} \in \mathbb{R}^{n\times n}$, $\boldsymbol{B} \in \mathbb{R}^{n\times p}$, $\boldsymbol{C} \in \mathbb{R}^{q\times n}$, $\boldsymbol{D} \in \mathbb{R}^{q\times p}$. In HiPPO~\cite{gu2020hippo} (high-order polynomial projection operator), continuous time memorisation is posed as a problem of function approximation in a Hilbert space defined by a probability measure $\mu$. For a \emph{scaled Legendre} probability measure, one obtains the HiPPO matrix $\boldsymbol{A}$, which enforces uniform weight in the memorisation of all previously observed inputs, in contrast to the exponentially decaying weighting of the constant error carousel of LSTMs~\cite{hochreiter1997long}. The HiPPO mode of memorisation is shown empirically to be better suited to modeling long-range dependencies (LRD) than other neural memory layers, for which it serves as a drop-in replacement.

Whereas in HiPPO, the state matrix $\boldsymbol{A}$ is a fixed constant, the linear state space layer (LSSL)~\cite{gu2021combining} incorporates $\boldsymbol{A}$ as a learnable parameter. However, this increased expressiveness introduces intractable powers of $\boldsymbol{A}$. In~\cite{gu2021efficiently}, the LSSL is instead reparameterised as the sum of diagonal and low-rank matrices, allowing for the efficient computation of the layer kernel in Fourier space. This updated formulation is known as the \emph{structured} state space sequence layer (S4). Note that as a linear operator, the inverse discrete Fourier transform is amenable to backpropagation in the context of a neural network. Note also that under this formulation, the hidden state $x(t)$ is only computed implicitly. Finally, \cite{gu2022parameterization} presents a simplification of the S4 layer, known as diagonal S4 (S4D), in which $\boldsymbol{A}$ is approximated by a diagonal matrix.

%% file: sections/3_methods.tex
% Formalise loss function, define dimensionality, $N$, sequence length $L$ etc.

% Preprocessing (splitting into patches, feature extraction)
% include diagram here

% Architecture

Given that the patch extraction of whole slide images at high magnifications results in long sequences of patches, we propose to incorporate a state space layer in a MIL aggregation network to better represent each patch sequence.

\subsection{Neural state space models}

In practice, neural state space models (SSM) simulate Equation \ref{eq:ss} in discrete time, invoking a recurrence relation on the discretised hidden state,

\begin{equation}
\label{eq:discetised_ss}
\begin{aligned}
x_t &= \overline{\boldsymbol{A}}x_{t-1} + \overline{\boldsymbol{B}}u_t \\ 
y_t &= \overline{\boldsymbol{C}}x_{t} + \overline{\boldsymbol{D}}u_t
\end{aligned}
\end{equation}

where the sequences $u_t$, $x_t$, and $y_t$ are the discretised $u(t)$, $x(t)$, and $y(t)$, and the modified model parameters arise from a bilinear discretisation~\cite{gu2021combining}. As such, SSMs bear an inherent resemblance to RNNs, where the hidden representation $x_{t}$ can be interpreted as a memory cell for the observed sequence over the interval $[0, t]$, and with $\overline{\boldsymbol{D}}u_t$ acting as a skip connection between the input and output at point $t$. Due to their lack of non-linearities, state space models can also be viewed as a convolution between two discrete sequences. Playing out the recurrence in Equation \ref{eq:discetised_ss}, one obtains,

\begin{equation}
\label{eq:vandermonde}
y = \overline{\boldsymbol{K}} \ast u + \overline{\boldsymbol{D}}u,
\end{equation}

where $u \in \mathbb{R}^L$ and $y \in \mathbb{R}^L$ are the full input and output sequences, and the sequence $\overline{\boldsymbol{K}} \in \mathbb{R}^L$ is defined as,

\begin{equation}
    \overline{\boldsymbol{K}} = (\overline{\boldsymbol{C}\boldsymbol{B}}, \overline{\boldsymbol{C}\boldsymbol{A}\boldsymbol{B}}, \dots, \overline{\boldsymbol{C}\boldsymbol{A}}^{L-1}\overline{\boldsymbol{B}}),
\end{equation}

% \begin{equation}
% \label{eq:discetised_ss}
% \begin{aligned}
% y &= \overline{\boldsymbol{K}} \ast u  \\ 
% \overline{\boldsymbol{K}} &= (\overline{\boldsymbol{C}\boldsymbol{B}}, \overline{\boldsymbol{C}\boldsymbol{A}\boldsymbol{B}}, \dots, \overline{\boldsymbol{C}\boldsymbol{A}^{L-1}\boldsymbol{B}})
% \end{aligned}
% \end{equation}

which is computed efficiently by the S4D algorithm~\cite{gu2022parameterization}. Note that although SSM layers are linear, they may be combined with other, non-linear layers in a neural network. Note also that although Equation~\ref{eq:vandermonde} is posed as modeling a one-dimensional signal, in practice multi-dimensional inputs are modelled simply by stacking SSM layers together, followed by an affine ``mixing'' layer.

\begin{figure}[t]
\centering
  \includegraphics[width=0.95\textwidth]{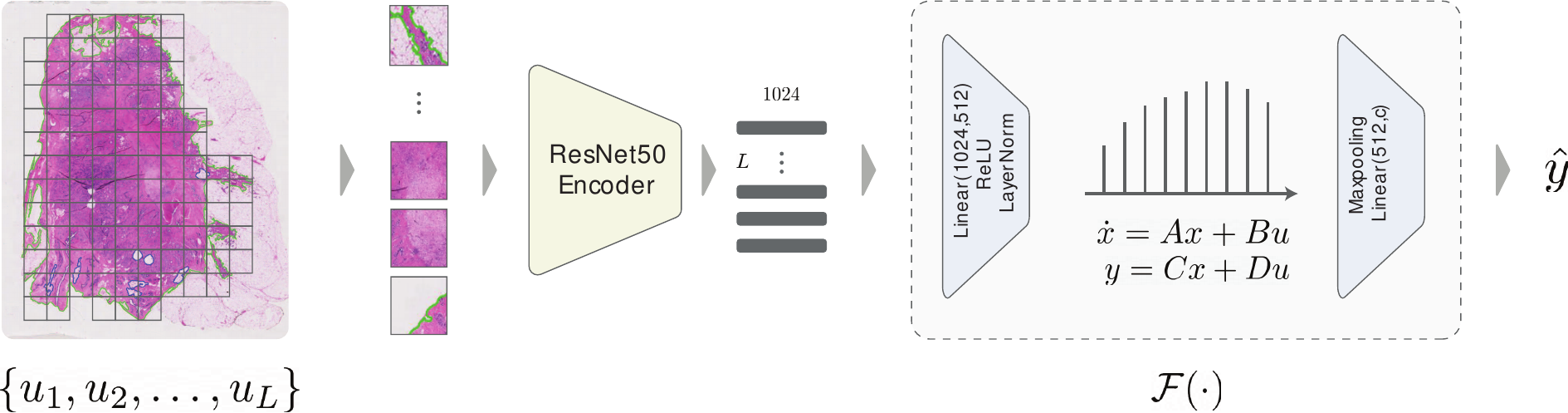}%     without .tex extension
  % or use \input{mytikz}
  \caption{Overview of the proposed pipeline. In the first step, patches are extracted from a regular grid on a WSI. These patches are embedded using a pre-trained ResNet50 and are aggregated by a sequence model based on a state space layer.}
  \label{fig:pipeline}
\end{figure}

\subsection{MIL training}

In our pipeline (Figure \ref{fig:pipeline}) WSIs are first divided into a sequence of $L$ patches $\{u_1, u_2, \dots, u_L\}$, where $L$ will vary by slide. A pretrained ResNet50 is then used to extract a 1024-dimensional feature vector from each patch $\{\mathbf{u}_1, \mathbf{u}_2, \dots, \mathbf{u}_L\}$, which constitute the model inputs. We define a SSM-based neural network $\mathcal F$ to predict a WSI-level class probability given this input sequence,

\begin{equation}
\hat{y} = \mathcal F(\{\mathbf{u}_1, \mathbf{u}_2, \dots, \mathbf{u}_L\}).
\end{equation}

The architecture of $\mathcal F$ is composed of an initial linear projection layer, used to lower the dimensionality of each vector in the input sequence. A SSM layer is then applied feature-wise by applying the S4D algorithm. That is, Equation~\ref{eq:vandermonde}, including the skip connection, transforms the sequence $\{u_{1,d}, u_{2,d}, \dots, u_{L,d}\}$ for all features $d$, and the resulting sequences are concatenated. A linear ``mixing'' layer is applied token-wise, doubling the dimensionality of each token, followed by a gated linear unit~\cite{dauphin2017language} acting as an output gate, which restores the input dimensionality. For the SSM layer, we used the official implementation of S4D\footnote{\href{https://github.com/HazyResearch/state-spaces}{https://github.com/HazyResearch/state-spaces}}. A max pooling layer merges the SSM layer outputs into a single vector, which is projected by a final linear layer and softmax to give the class probabilities $\hat{y}$. The model is trained according to,

\begin{equation}
\label{eq:losloss}
\mathcal{L}_{MIL} = -\frac{1}{M}\sum_{m=1}^M \log \hat{y}_{c_m},
\end{equation}

where $\hat{y}_{c_m}$ denotes the probability corresponding to $c_m$, the slide-level label of the sequence corresponding to the $m^{\text{th}}$ of $M$ whole slide images.

\subsection{Multitask training}

One advantage of processing an entire slide as a sequence is the ease with which additional supervision may be incorporated, when available.
%at the patch level. Some datasets such as CAMELYON16 offer both a slide-level label indicating patient diagnosis, as well as patch-level labels deriving from expert annotations of metastatic regions. 
A patch-level ground truth creates the opportunity for multitask learning, which can enhance the representations learned for slide-level classification. As an extension of our base model in Equation \ref{eq:losloss}, we train a multitask model to jointly predict a slide-level and patch-level labels. Prior to the max pooling layer of the base model, an additional linear layer is applied to each sequence token, yielding $L$ additional model outputs. This multitask model is trained according to a sum of log losses,

\begin{equation}
\mathcal{L}_{MT} = -\frac{1}{M}\sum_{m=1}^M\bigg( \log \hat{y}_{c_m} + \frac{\lambda}{L} \cdot \sum_{l=1}^L \log \hat{y}_{c_{m,l}} \bigg),
\label{eq:multitask}
\end{equation}

where $c_{m,l}$ indexes the class of the $l^{\text{th}}$ patch in the $m^{\text{th}}$ training slide and $\lambda$ is a tunable hyperparameter used to modulate the relative importance of each task.

\subsection{Implementation details}

We extracted patches of size $256\times256$ from the tissue regions of WSIs at $20$x magnification. Following CLAM~\cite{lu2021data}, the third residual block of a pretrained ResNet50~\cite{he2016deep} was used as a feature extractor, followed by a mean pooling operation, resulting in a $1024$-dimensional representation for each patch. These features were used as inputs to all models.
All model training was performed under a $10$-fold cross-validation, and all reported results are averaged over the validation sets of the folds, aside from CAMELYON16, for which the predefined test set was utilized. Thus, for CAMELYON16, we report test set performances averaged over the validation.

%Each slide is therefore described by a sequence of $L$ vectors of dimension 1024, with a varying sequence length, depending on the size of the tissue region. Given an sequence of input patches $\{x_1, ..., x_L\}$, each slide is therefore described by a sequence $\{\tilde{x}_1, ..., \tilde{x}_L\}$, with $\tilde{x}_i \in \mathbb R^{1024}$.
 
Baseline models were chosen to be prior art CLAM~\cite{lu2021data} and TransMIL~\cite{shao2021transmil}.  The official code of these two models was used to perform the comparison. In addition, we included a vanilla transformer, a LSTM RNN, and models based on mean and max pooling.
Our vanilla transformer is composed of two stacked self-attention blocks, with four attention heads, a model dimension of $256$, and a hidden dimension of $256$. For the LSTM, we used an embedding size of 256 and a width of $256$.
The pooling models applied pooling feature-wise across each sequence, then used a random forest with $200$ trees for classification. 
For the S4 models, the dimension of the state matrix $\boldsymbol{A}$ was tuned to $32$ for CAMELYON16 and TCGA-RCC, and $128$ for TCGA-LUAD. Our models were trained using the Adam~\cite{kingma2014adam} optimizer with the lookahead method~\cite{zhang2019lookahead}, with a learning rate of $2\cdot10^{-4}$, and weight decay of $10^{-4}$ for TCGA-LUAD and TCGA-RCC and $10^{-3}$ for CAMELYON16. Early stopping with a patience of $10$ was used for all our training. Our implementation is publicly available\footnote{\href{https://github.com/MICS-Lab/s4\_digital\_pathology}{https://github.com/MICS-Lab/s4\_digital\_pathology}}.

%The tissue region is determined using Otsu thresholding~\cite{otsu1979threshold}.
%Feature vectors of dimension 1024 are then extracted from each patch using by averaging the activation maps from the third residual block of a  ResNet50~\cite{he2016deep} feature extractor as performed in the CLAM~\cite{lu2021data} pipeline. Each slide is therefore described by a sequence of $L$ vectors of dimension 1024, with a varying sequence length, depending on the size of the tissue region. Given an sequence of input patches $\{x_1, ..., x_L\}$, each slide is therefore described by a sequence $\{\tilde{x}_1, ..., \tilde{x}_L\}$, with $\tilde{x}_i \in \mathbb R^{1024}$.

%% file: sections/4_results.tex
\subsection{Data}
% Datasets (describe the type of data, the centers it comes from, the magnification at which it is extracted, the number of slides, the sequence lengths)
\textbf{CAMELYON16}~\cite{litjens20181399} is a dataset that consists of resections of lymph nodes, where each WSI is annotated with a binary label indicating the presence of tumour tissue in the slide, and all slides containing tumors have a pixel-level annotation indicating the metastatic region. In multitask experiments, we use this annotation to give each patch a label indicating local tumour presence. There are 270 WSIs in the training/validation set, and 130 WSIs in the predefined test set. In our experiments, the average patch sequence length arising from CAMELYON16 is 6129 (ranging from 127 to 27444).

\textbf{TCGA-LUAD} is a TCGA lung adenocarcinoma dataset that contains 541 WSIs along with genetic information about each patient. We obtained genetic information for this cohort using Xena browser~\cite{goldman2020visualizing}. As a MIL task, we chose the task of predicting the patient mutation status of TP53, a tumor suppressor gene that is highly relevant in oncology studies. The average sequence length is 10557 (ranging from 85 to 34560).

\textbf{TCGA-RCC} is a TCGA dataset for three kidney cancer subtypes (denoted KICH, KIRC, and KIRP). It consists of 936 WSIs (121 KICH, 518 KIRC, and 297 KIRP). The average sequence length is 12234 (ranging from 319 to 62235).

\subsection{Results}
\textbf{Multiple instance learning results.} We evaluate our method on each dataset by accuracy and area under receiver operating characteristic curve (AUROC). For multiclass classification, these were computed in a one-versus-rest manner.

Table~\ref{tab:main_results} summarises the comparison between our proposed model and baselines. For the CAMELYON16 dataset, our method performs on par with TransMIL and the CLAM models, while it clearly outperforms the other methods. Similarly, in the TCGA-LUAD dataset the proposed model achieves comparable performance with both CLAM models, while outperforming TransMIL and the other methods. We note that TCGA-LUAD proves to be a more challenging dataset for all models. Moreover, our method outperforms CLAM models on the TCGA-RCC dataset, while reporting very similar performance with respect to TransMIL. Overall, looking at the average metrics per model across all three datasets, our proposed method achieves the highest accuracy and the second highest AUROC, only behind CLAM-MB. A pairwise t-test between the proposed method, CLAM, and TransMIL shows that there is no statistical significance performance difference (see supplementary material). 

%Camelyon
%The task of binary classification on the Camelyon16 dataset shows that btoh CLAM models, TransMIL, and our models score similarly and clearly outperform all other methods in terms of accuracy. However, TransMIL achieves higher AUROC than other methods.

%TCGA-LUAD
%The TCGA-LUAD TP53 classification is a difficult task, which no current model is considered to have solved. Neither the pooling-based models or the LSTM and Transformer models manage to get high results. In terms of AUROC, both CLAM models seem to outperform other methods, but regarding the accuracy, both the CLAM models and our approach seem to score similarly and to outperform all other methods.

%TCGA-RCC
%On the TCGA-RCC dataset, it is interesting to see how well the max-pooling model is able to perform, even better than models such as LSTMs or Transformers which perhaps fail to model such long sequences. CLAM models are able to outperform the max-pooling in terms of AUROC, but score similarly in terms of accuracy. Both TransMIL and our model score similarly on both metrics and outperform all other tested methods.

% MAIN RESULTS TABLE
\begin{table}[t!]
\centering
\caption{Comparison of accuracy and AUROC on three datasets CAMELYON16, TCGA-LUAD, TCGA-RCC, and on average. All metrics in the table are the
average of 10 runs. Best performing methods are indicated in \textbf{bold} and second best in \textit{italics}. $^\ast$ indicates results from \cite{shao2021transmil}.}
\label{tab:main_results}
\begin{tabular}{c c c c c c c c c}
\toprule
\multicolumn{1}{c}{Dataset} & \multicolumn{2}{c}{CAMELYON16} & \multicolumn{2}{c}{TCGA-LUAD} &
\multicolumn{2}{c}{TCGA-RCC} & \multicolumn{2}{c}{Average}\\
% \midrule
Metric & \multicolumn{1}{c}{Acc.} & \multicolumn{1}{c}{AUROC} & \multicolumn{1}{c}{Acc.} & \multicolumn{1}{c}{AUROC} & \multicolumn{1}{c}{Acc.} & \multicolumn{1}{c}{AUROC} & \multicolumn{1}{c}{Acc.} & \multicolumn{1}{c}{AUROC} \\

\midrule
Mean-pooling    & 0.5969 & 0.5810 &     0.6261 & 0.6735 &       0.8608 & 0.9612 & 0.6946 & 0.7386 \\
Max-pooling     & 0.7078 & 0.7205 &     0.6328 & 0.6686 &       0.8803 & 0.9659 & 0.7403 & 0.7850\\
Transformer \cite{vaswani2017attention}
& 0.5419 & 0.5202 &     0.5774 & 0.6214 &       0.7932 & 0.9147 & 0.6375 & 0.6854\\
LSTM \cite{graves2012long}
& 0.5310 & 0.5053 &     0.5389 & 0.5208 &       0.6654 & 0.7853 & 0.5784 & 0.6038\\
CLAM SB \cite{lu2021data}
& 0.8147 & 0.8382 &     0.6859 & \textit{0.7459} &       $0.8816^\ast$ & $0.9723^\ast$ & 0.7941 & 0.8532\\
CLAM MB \cite{lu2021data}
& \textit{0.8264} & \textit{0.8523} &     \textbf{0.6901} & \textbf{0.7573} &       $0.8966^\ast$ & $0.9799^\ast$ & \textit{0.8044} & \textbf{0.8632}\\
TransMIL \cite{shao2021transmil}    
& \textbf{0.8287} & \textbf{0.8628} &     0.6348 & 0.7015 &       \textbf{0.9466}$^\ast$ & \textit{0.9882}$^\ast$  & 0.8034 & 0.8508\\
\midrule
Ours
& 0.8217 & 0.8485 &     \textit{0.6879} & 0.7304 &       \textit{0.9426} & \textbf{0.9885}  & \textbf{0.8174} & \textit{0.8558}\\
\bottomrule
\end{tabular}
\end{table}

We further compare our method with respect to model and time complexity. In Table \ref{tab:param_results} we report the number of trainable parameters, as well as the inference time for all models. The number of parameters is computed with all models configured to be binary classifiers, and the inference time is computed as the average time over $100$ samples for processing a random sequence of $1024$-dimensional vectors of length $30000$. For our proposed method, we report both models with the different state dimensions (Ours ($SSM_{32}$)) and (Ours ($SSM_{128}$)). Compared with TransMIL, our method runs four times faster and has less than half the parameters. The CLAM models are more efficient in terms of number of trainable parameters, yet CLAM MB is slower.

% PARAMETERS AND INFERENCE TIME TABLE
% Note : the number of parameters is for a binary classification model
%        the inference time is the mean (N=100) inference time for random input of (1, 30000, 1024)
\begin{table}[t!]
\centering
\caption{Comparison of parameter count and inference time for all methods.}
\label{tab:param_results}
\begin{tabular}{*3c}
\toprule
Model & Number of parameters & Inference time (ms)\\
\midrule
Mean-pooling & 1 025 & 5.60\\
Max-pooling & 1 025 & 77.49\\
Transformer \cite{vaswani2017attention} & 1 054 978 & 2.60\\
LSTM \cite{graves2012long} & 789 250 & 320.52\\
CLAM SB \cite{lu2021data} & 790 791 & 0.84\\
CLAM MB \cite{lu2021data} & 791 048 & 5.85\\
TransMIL \cite{shao2021transmil} & 2 672 146 & 8.58\\
\midrule
Ours (SSM$_{128}$) & 1 184 258 & 2.01\\
Ours (SSM$_{32}$) & 1 085 954 & 1.97\\
\bottomrule
\end{tabular}
\end{table}

Table \ref{tab:ablations_results} shows the effect of modifying parts of the architecture on the results for TCGA-RCC. Most modifications had very little impact on AUROC, but a more significant impact can be seen on the accuracy of the model.
%In model A, we added a positional encoding, which resulted in a significant drop of accuracy. It can be hypothesised that for the tasks of classification, the relative two-dimensional position of the patch did not bring any additional information. It is possible that on other tasks adding positional encoding will bring relevant information.
Models A and B show that stacking multiple SSM layers results in lower accuracy, which was observed over all three datasets, while models C and D show that modifying the state dimension of the SSM module can have an impact on the accuracy. The optimal state space dimension varies depending on the dataset.

\begin{table}
\centering
\caption{Ablation study for the different SSM components on the TCGA-RCC dataset. Best results in \textbf{bold}.}
\label{tab:ablations_results}
\begin{tabular}{*5c}
\toprule
Model & SSM layers & State dimension & Accuracy & AUROC\\
\midrule
A & 2 & 32 & 0.9236 & 0.9813\\
B & 3 & 32 & 0.9179 & 0.9834\\
C & 1 & 16 & 0.9352 & 0.9846\\
D & 1 & 64 & 0.9352 & 0.9861\\
\midrule
Ours & 1 & 32 & \textbf{0.9426} & \textbf{0.9885}\\
\bottomrule
\end{tabular}
\end{table}

\textbf{Multitask learning results.} We explored the ability of our model to combine slide- and patch-level information on the CAMEYLON16 dataset. We compared our model with the best performing model on CAMELYON16, TransMIL. Both models were trained according to Equation \ref{eq:multitask} with $\lambda=5$ tuned by hand. In Table \ref{table:multitask} we give slide-level accuracy and AUROC for the two models. We observe that all accuracies and AUROC increase compared with those reported in Table \ref{tab:main_results}. This indicates that the use of patch-level annotations complements the learning of the slide-level label. We furthermore observe that our model outperforms TransMIL when combining slide- and patch-level annotations. We map the sequence of output probabilities to their slide coordinates giving a heatmap localising metastasis (see supplementary material).

\begin{table}[t!]
\centering
\caption{Comparison of accuracy and AUROC for models trained as multitask classifiers on the CAMELYON16 dataset. Best results in \textbf{bold}.}
\label{tab:multitask_results}
\begin{tabular}{*3c}
\toprule
Model & Accuracy & AUROC\\
\midrule
TransMIL~\cite{shao2021transmil} & $0.8403$ & $0.8828$\\
Ours & $\mathbf{0.8488}$ & $\mathbf{0.8998}$\\
\bottomrule
\end{tabular}
\label{table:multitask}
\end{table}

\textbf{Performance on longest sequences.} In order to highlight the inherent ability of SSM models to effectively model long sequences, we performed an experiment on only the largest WSIs of the TCGA-RCC dataset. Indeed, this dataset contains particularly long sequences (up to $62235$ patches at $20$x). We evaluated the trained models for each fold on a subset of the validation set, only containing sequences with a length in the 85$^\text{th}$ percentile. Table \ref{tab:long_sequences_results} shows the obtained average accuracy (weighted by the number of long sequences in each validation set) and AUROC on both CLAM models, TransMIL, and our proposed method. Both in terms of AUROC and accuracy, our method outperforms the other methods on long sequences, while the performances are comparable to Table \ref{tab:main_results}, albeit slightly lower, illustrating the challenge of processing large WSIs.

%+ Why RCC, what was the constraint exactly. And a shord discusion on the result.

\begin{table}[t!]
\centering
\caption{Results of CLAM SB, CLAM MB, TransMIL, and our proposed method on long sequences. Best results in \textbf{bold}.}
\label{tab:long_sequences_results}
\begin{tabular}{*3c}
\toprule
Model & Accuracy & AUROC\\
\midrule
CLAM SB \cite{lu2021data} & 0.9149 & 0.9635\\
CLAM MB \cite{lu2021data} & 0.8936 & 0.9654\\
TransMIL~\cite{shao2021transmil} & 0.9007 & 0.9652\\
Ours & \textbf{0.9220} & \textbf{0.9737}\\
\bottomrule
\end{tabular}
\end{table}

%% file: sections/5_conclusions.tex
In this work we have explored the ability of state space models to act as multiple instance learners on sequences of patches extracted from histopathology images. These models have been developed for their ability to memorise long sequences, and they have proven competitive with state of the art MIL models across a range of pathology problems. Additionally, we demonstrated the ability of these models to perform multiclass classification, which furthermore allowed us to visualise the localisation of metastasic regions. Finally, we demonstrated that on the longest sequences in our datasets, state space models offer better performance than competing models, confirming their power in modeling long-range dependencies.

% Maybe a comment on the limitation of the attention extraction. Probably with a promise for an extension are future work.

\subsubsection{Acknowledgments}
This work has benefited from state financial aid, managed by the Agence Nationale de Recherche under the investment program integrated into France 2030, project reference ANR-21-RHUS-0003. This work was partially supported by the ANR  Hagnodice ANR-21-CE45-0007. Experiments have been conducted using HPC resources from the \href{http://mesocentre.centralesupelec.fr/}{“Mésocentre”} computing center of CentraleSupélec and École Normale Supérieure Paris-Saclay supported by CNRS and Région Île-de-France.

%% file: sections/6_supplementary.tex
\begin{figure}
\centering
\subfloat[TransMIL]{\includegraphics[width=0.32\textwidth]{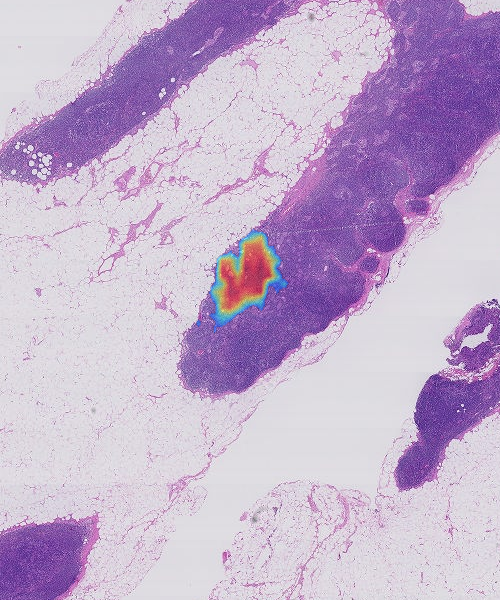}}\label{subfig:ihcsamples}
\subfloat[Ours]{\includegraphics[width=0.32\textwidth]{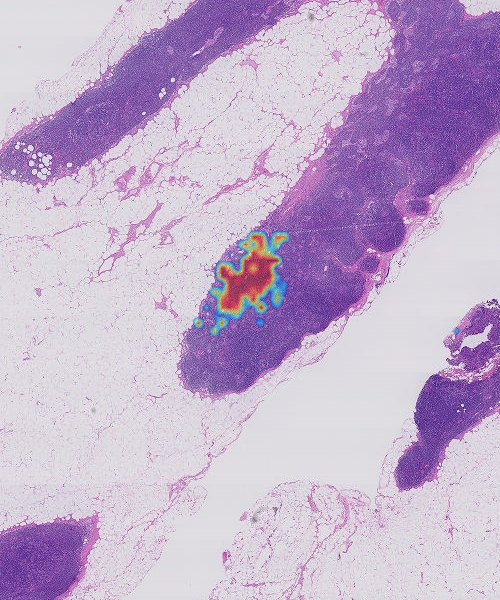}}\label{subfig:ihcsamples}
\subfloat[Ground truth mask]{\includegraphics[width=0.32\textwidth]{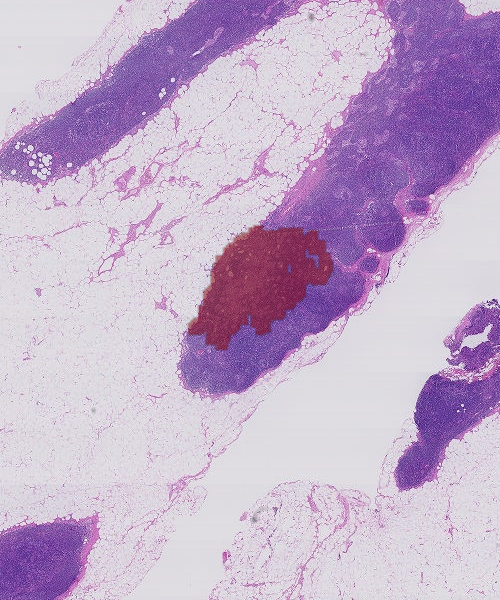}}\label{subfig:hessamples} 
\caption{Heatmap visualisations of metastatic region after multitask training on CAMELYON16 for TransMIL (a), proposed model (b), and ground truth (c).}
\label{fig:heatmaps}
\end{figure}

\begin{figure}
\centering
\includegraphics[width=0.99\textwidth]{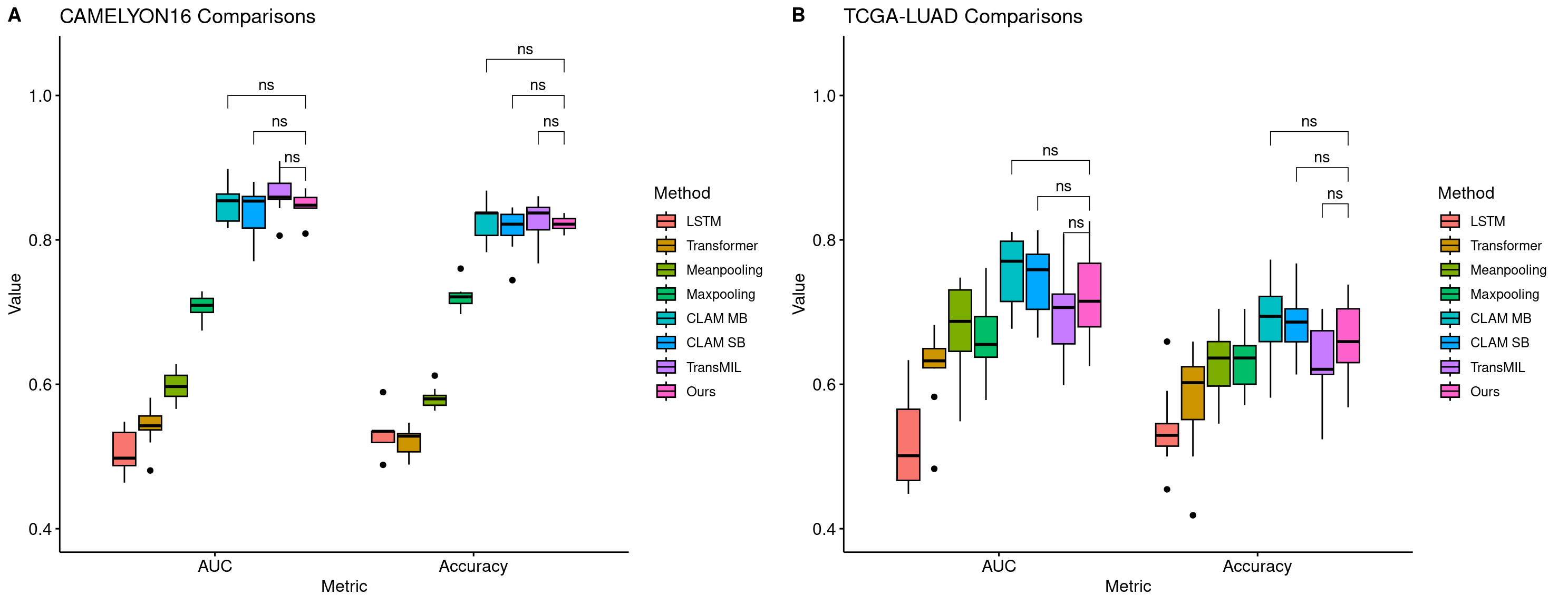}
\caption{AUC and accuracy over 10 folds for all methods on the CAMELYON16 dataset (A) and the TCGA-LUAD dataset (B). Significance estimated with pairwise $t$-tests (confidence level of 0.95)}
\label{fig:suppl_1}
\end{figure}

%% file: main.bbl
\begin{thebibliography}{10}
\providecommand{\url}[1]{\texttt{#1}}
\providecommand{\urlprefix}{URL }
\providecommand{\doi}[1]{https://doi.org/#1}

\bibitem{agarwalla2017representation}
Agarwalla, A., Shaban, M., Rajpoot, N.M.: Representation-aggregation networks
  for segmentation of multi-gigapixel histology images. BMVC  (2017)

\bibitem{campanella2019clinical}
Campanella, G., Hanna, M.G., Geneslaw, L., Miraflor, A., Silva, V.W.K., Busam,
  K.J., Brogi, E., Reuter, V.E., Klimstra, D.S., Fuchs, T.J.: Clinical-grade
  computational pathology using weakly supervised deep learning on whole slide
  images. Nature medicine  \textbf{25}(8),  1301--1309 (2019)

\bibitem{chen2022scaling}
Chen, R.J., Chen, C., Li, Y., Chen, T.Y., Trister, A.D., Krishnan, R.G.,
  Mahmood, F.: Scaling vision transformers to gigapixel images via hierarchical
  self-supervised learning. In: Proceedings of the IEEE/CVF Conference on
  Computer Vision and Pattern Recognition. pp. 16144--16155 (2022)

\bibitem{coudray2018classification}
Coudray, N., Ocampo, P.S., Sakellaropoulos, T., Narula, N., Snuderl, M.,
  Feny{\"o}, D., Moreira, A.L., Razavian, N., Tsirigos, A.: Classification and
  mutation prediction from non--small cell lung cancer histopathology images
  using deep learning. Nature medicine  \textbf{24}(10),  1559--1567 (2018)

\bibitem{dauphin2017language}
Dauphin, Y.N., Fan, A., Auli, M., Grangier, D.: Language modeling with gated
  convolutional networks. In: International conference on machine learning. pp.
  933--941. PMLR (2017)

\bibitem{gao2023semi}
Gao, Z., Hong, B., Li, Y., Zhang, X., Wu, J., Wang, C., Zhang, X., Gong, T.,
  Zheng, Y., Meng, D., et~al.: A semi-supervised multi-task learning framework
  for cancer classification with weak annotation in whole-slide images. Medical
  Image Analysis  \textbf{83},  102652 (2023)

\bibitem{goldman2020visualizing}
Goldman, M.J., Craft, B., Hastie, M., Repe{\v{c}}ka, K., McDade, F., Kamath,
  A., Banerjee, A., Luo, Y., Rogers, D., Brooks, A.N., et~al.: Visualizing and
  interpreting cancer genomics data via the xena platform. Nature biotechnology
   \textbf{38}(6),  675--678 (2020)

\bibitem{graves2012long}
Graves, A.: Long short-term memory. Supervised sequence labelling with
  recurrent neural networks pp. 37--45 (2012)

\bibitem{gu2020hippo}
Gu, A., Dao, T., Ermon, S., Rudra, A., R{\'e}, C.: Hippo: Recurrent memory with
  optimal polynomial projections. Advances in neural information processing
  systems  \textbf{33},  1474--1487 (2020)

\bibitem{gu2021efficiently}
Gu, A., Goel, K., R{\'e}, C.: Efficiently modeling long sequences with
  structured state spaces. arXiv preprint arXiv:2111.00396  (2021)

\bibitem{gu2022parameterization}
Gu, A., Gupta, A., Goel, K., R{\'e}, C.: On the parameterization and
  initialization of diagonal state space models. arXiv preprint
  arXiv:2206.11893  (2022)

\bibitem{gu2021combining}
Gu, A., Johnson, I., Goel, K., Saab, K., Dao, T., Rudra, A., R{\'e}, C.:
  Combining recurrent, convolutional, and continuous-time models with linear
  state space layers. Advances in neural information processing systems
  \textbf{34},  572--585 (2021)

\bibitem{he2016deep}
He, K., Zhang, X., Ren, S., Sun, J.: Deep residual learning for image
  recognition. In: Proceedings of the IEEE conference on computer vision and
  pattern recognition. pp. 770--778 (2016)

\bibitem{hochreiter1997long}
Hochreiter, S., Schmidhuber, J.: Long short-term memory. Neural computation
  \textbf{9}(8),  1735--1780 (1997)

\bibitem{kingma2014adam}
Kingma, D.P., Ba, J.: Adam: A method for stochastic optimization. arXiv
  preprint arXiv:1412.6980  (2014)

\bibitem{litjens20181399}
Litjens, G., Bandi, P., Ehteshami~Bejnordi, B., Geessink, O., Balkenhol, M.,
  Bult, P., Halilovic, A., Hermsen, M., van~de Loo, R., Vogels, R., et~al.:
  1399 h\&e-stained sentinel lymph node sections of breast cancer patients: the
  camelyon dataset. GigaScience  \textbf{7}(6),  giy065 (2018)

\bibitem{lu2021data}
Lu, M.Y., Williamson, D.F., Chen, T.Y., Chen, R.J., Barbieri, M., Mahmood, F.:
  Data-efficient and weakly supervised computational pathology on whole-slide
  images. Nature biomedical engineering  \textbf{5}(6),  555--570 (2021)

\bibitem{russakovsky2015imagenet}
Russakovsky, O., Deng, J., Su, H., Krause, J., Satheesh, S., Ma, S., Huang, Z.,
  Karpathy, A., Khosla, A., Bernstein, M., et~al.: Imagenet large scale visual
  recognition challenge. International Journal of Computer Vision
  \textbf{115}(3),  211--252 (2015)

\bibitem{shao2021transmil}
Shao, Z., Bian, H., Chen, Y., Wang, Y., Zhang, J., Ji, X., et~al.: Transmil:
  Transformer based correlated multiple instance learning for whole slide image
  classification. Advances in neural information processing systems
  \textbf{34},  2136--2147 (2021)

\bibitem{tellez2019neural}
Tellez, D., Litjens, G., van~der Laak, J., Ciompi, F.: Neural image compression
  for gigapixel histopathology image analysis. IEEE transactions on pattern
  analysis and machine intelligence  \textbf{43}(2),  567--578 (2019)

\bibitem{vaswani2017attention}
Vaswani, A., Shazeer, N., Parmar, N., Uszkoreit, J., Jones, L., Gomez, A.N.,
  Kaiser, {\L}., Polosukhin, I.: Attention is all you need. Advances in neural
  information processing systems  \textbf{30} (2017)

\bibitem{wang2020defense}
Wang, K., Oramas, J., Tuytelaars, T.: In defense of lstms for addressing
  multiple instance learning problems. In: Proceedings of the Asian Conference
  on Computer Vision (2020)

\bibitem{zhang2019lookahead}
Zhang, M.R., Lucas, J., Hinton, G., Ba, J.: Lookahead optimizer: K steps
  forward, 1 step back. In: Proceedings of the 33rd International Conference on
  Neural Information Processing Systems. Curran Associates Inc., Red Hook, NY,
  USA (2019)

\end{thebibliography}
